\documentclass[11pt,a4paper]{article}
\usepackage[hyperref]{emnlp-ijcnlp-2019}
\usepackage{times}
\usepackage{latexsym}
\usepackage{tabularx}
\usepackage{amssymb}
\usepackage{makecell}
\usepackage{float}
\usepackage{comment}
\usepackage{graphicx}
\usepackage{multirow}
\usepackage{multicol}
\usepackage{textcomp}
\usepackage[table-number-alignment=center, group-separator={}]{siunitx}
\def\sym#1{\ifmmode^{#1}\else\(^{#1}\)\fi}
\usepackage{url}
\usepackage{booktabs}
\usepackage{hyperref}
\newcommand{\Sref}[1]{Appendix~\ref{#1}}

\aclfinalcopy %

\newcommand\rquoted{\textsc{Quoted}}
\newcommand\rcontext{\textsc{Context}}
\newcommand\rcomment{\textsc{Comment}}
\newcommand\rreply{\textsc{Reply}}
\newcommand\bertb{\emph{BERT}\ensuremath{_B}}
\newcommand\bertl{\emph{BERT}\ensuremath{_L}}
\newcommand{\datasetname}{TalkDown}
\newcommand{\dataset}{\textsc{\datasetname}}

\makeatletter
\newcommand\footnoteref[1]{\protected@xdef\@thefnmark{\ref{#1}}\@footnotemark}
\makeatother

\title{\datasetname: A Corpus for Condescension Detection in Context}

\author{Zijian Wang \\
  Symbolic Systems Program\\ %
  Stanford University \\
  {\tt zijwang@stanford.edu} \\\And
  Christopher Potts \\
  Department of Linguistics \\
  Stanford University \\
  {\tt cgpotts@stanford.edu} \\}

\date{}

\begin{document}
\maketitle
\begin{abstract}
Condescending language use is caustic; it can bring dialogues to an end and bifurcate communities.
Thus, systems for condescension detection could have a large positive impact.
A challenge here is that condescension is often impossible to detect from isolated utterances, as it depends on the discourse and social context. To address this, we present \dataset, a new labeled dataset of condescending
linguistic acts in context. We show that extending a language-only model with
representations of the discourse improves performance, and we motivate techniques for dealing with the low rates of condescension overall.
We also use our model to estimate condescension rates in various online communities and relate these differences to differing community norms.
\end{abstract}

\section{Introduction}
Condescending language use can derail conversations and, over time,
disrupt healthy communities. The caustic nature of this language
traces in part to the ways that it keys into differing social roles
and levels of power \citep{fournier2002social}. It is common for
people to be condescending without realizing it \cite{wong2014and},
but a lack of intent only partly mitigates the damage it can cause. Thus, condescension detection is a
potentially high-impact NLP task that could open the door to many applications and future research directions, including, for example, supporting productive interventions in online communities \citep{Spertus:1997}, educating people who use condescending language in writing, helping linguists to understand the implicit linguistic acts associated with condescension, and helping social scientists to study the relationship between condescension and other variables like gender or socioeconomic status.

Progress on this task is currently limited by a lack of high-quality
labeled data. A deeper challenge is that condescension is often
impossible to detect from isolated utterances. First, a characteristic
of condescending language is that it is not overtly negative or
critical -- it might even include (insincere) praise
\cite{huckin2002critical}. Second, condescension tends to rest on a
pair of conflicting pragmatic presuppositions: a speaker presumption
that the speaker has higher social status than the listener, and a
listener presumption that this is incorrect. For example, an utterance
that is entirely friendly if said by one friend to another might be
perceived as highly condescending if said by a customer to a store
clerk. In such cases, the social roles of the participants shape the
language in particular ways to yield two very different outcomes.

\begin{figure}[tp]
  \centering
  \includegraphics[width=1\linewidth]{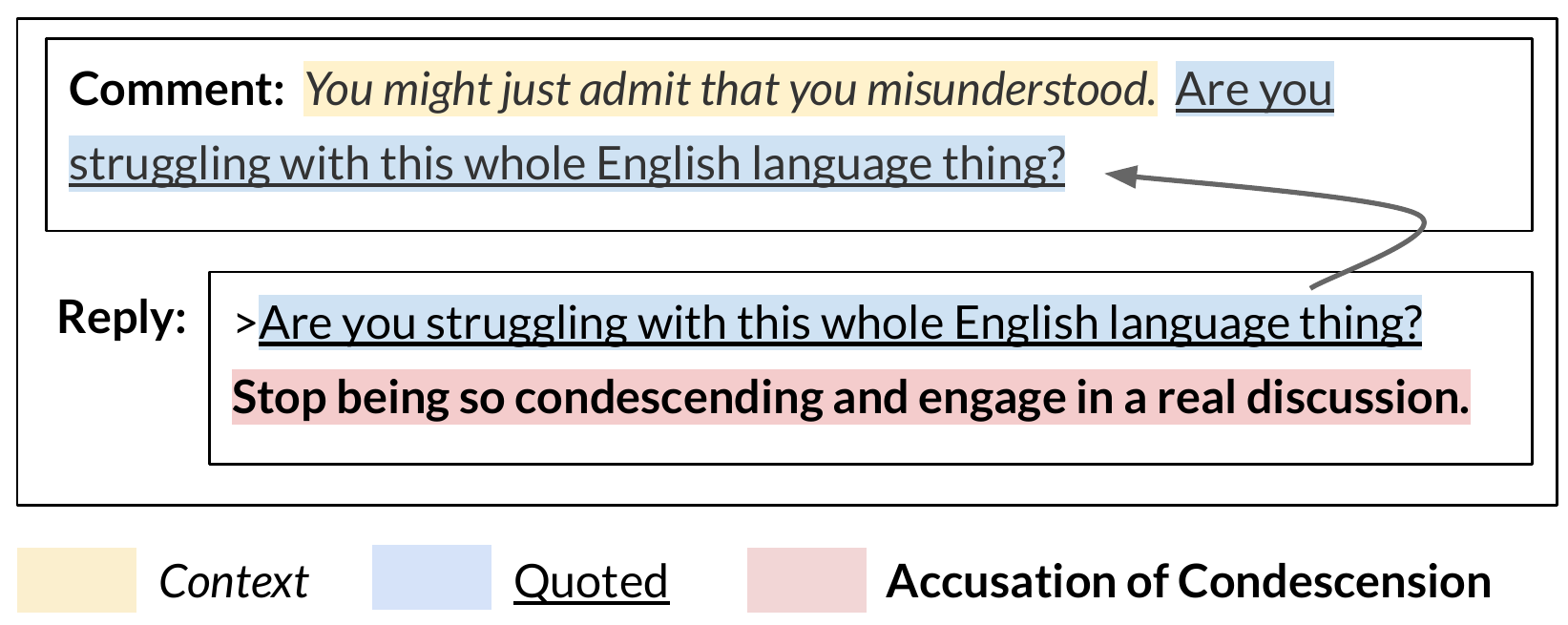}
  \caption{In this example, the \rreply{} quotes from part of
    the \rcomment{} and says that this \textsc{Quoted} text
    is condescending.} %
  \label{fig:data-example}
\end{figure}

In this paper, we seek to facilitate the development of models for
condescension detection by introducing \dataset, a new labeled dataset of
condescending acts in context. The dataset is derived from Reddit, a
thriving set of online communities that is diverse in content and
tone. We focus on \rcomment{} and \rreply{} pairs of the sort given in
Figure~\ref{fig:data-example}, in which the \rreply{} targets a
specific quoted span (\rquoted{}) in the \rcomment{} as being condescending. The
examples were multiply-labeled by crowdsourced workers,
which ensures high-quality labels and allows us to include
nuanced examples that require human judgment.

Our central hypothesis is that context is decisive for condescension
detection. To test this, we evaluate models that seek to make
this classification based only on the \rquoted{} span in the \rcomment{} as well as extensions of those models
that include summary representations of the preceding linguistic
context, which we treat as an approximation of the discourse context
in which the condescension accusation was made. Models with contextual
representations are far superior, bolstering the original
hypothesis. In addition, we show that these models are robust to
highly imbalanced testing scenarios that approximate the true low rate
of condescension in the wider world. Such robustness to imbalanced
data is an important prerequisite for deploying models like this.
Finally, we apply our model to a wide range of subreddits, arguing
that our estimated rates of condescension are related to different
community norms.

\section{The \textsc{\datasetname} Corpus}
We chose Reddit as the basis for our corpus for a few key reasons.
First, it is a large, publicly available dataset from an active set of more than one million user-created online communities (subreddits).\footnote{\scriptsize\url{http://redditmetrics.com/history}} Second, it varies in both content and
tone.
Third, users can develop strong identities on the
site, which could facilitate user-level modeling, but these identities
are generally pseudonymous, which is useful when studying charged
social phenomena  \citep{hamilton2017loyalty,wang2018s}. Fourth, the subreddit structure of the site creates
opportunities to study the impact of condescension on community
structure and norms \citep{buntain2014identifying,lin2017better,zhang2017community,chandrasekharan2018internet}.

The basis for our work is the Reddit data dump
2006--2018.\footnote{\scriptsize{\url{https://files.pushshift.io/reddit}}} We
first extracted \rcomment{}/\rreply{} pairs in which the \rreply{} contains a
condescension-related word. After further filtering out self-replies
and moderator posts, and normalizing links and references, we obtain
2.62M \rcomment{}/\rreply{} pairs.

Not all of these examples truly involve the \rreply{} saying that the
\textsc{Quoted} span is condescending. Our simple pattern-based
extraction method is not sufficiently sensitive. To address this, we
conducted an annotation project on Amazon Mechanical Turk. Our own
initial assessment of 200 examples using a five-point Likert scale
revealed two things that informed this project.

First, we saw a clear split in the positive instances of
condescension. In some, specific linguistic acts are labeled as
condescending (``This is really condesending''), whereas others
involve general user-level accusations that are not tied to specific
acts (``You're so condescending''). We chose to focus on the specific
linguistic acts. They provide a more objective basis for annotation,
and they can presumably be aggregated to provide an empirically
grounded picture of user-level behavior (or others' reactions to such
behavior). Thus, for positive instances of condescension, we further
limited our attention to \rcomment{}/\rreply{} pairs in which the
\rreply{} contains a direct quotation from the \rcomment, using fuzzy
match based on Levenshtein distance \cite{navarro2001guided}, as
illustrated in Figure~\ref{fig:data-example}. We extracted 66K such
examples. Some statistics on these examples is given in
Table~\ref{tab:corpus-stats}.

Second, with the above ambiguity addressed, the signal of condescending
or not is mostly clear. Thus, we designed the annotation project
around a three-way multiple choice question: \emph{condescending},
\emph{not condescending}, and \emph{cannot decide}. Each task began
with instructions and two training questions,
following by 10 different \rcomment{}/\rreply{} pairs to be labeled.
\Sref{appendix-data-annotation} provides screenshots of the annotation
interface.

\begin{table}[tp]
  \centering
  \resizebox{1\linewidth}{!}{
  \begin{tabular}[t]{r r r r r}
    \toprule
    & Median & Mean & Std. & Max \\
    \midrule
    \rquoted{} & 18 & 22.79 & 18.40 & 399 \\
    \rreply{}       & 67 & 100.92 & 112.34 & 1,921 \\
    \rcontext{}     & 47 & 98.40 & 154.59 & 2,136 \\
    \bottomrule
  \end{tabular}
  }
  \caption{Basic statistics of the length of the examples in the corpus.
    \rcontext{} is everything in the \rcomment{} before
    the \rquoted{} span.
  }
  \label{tab:corpus-stats}
\end{table}

To process the annotations, we filtered out the work of annotators who
did not correctly answer the training questions. The
remaining annotators have moderate to substantial agreement (Fleiss
$\kappa = 0.593$; \citealt{fleiss1971measuring,landis1977measurement}). We
then used Expectation--Maximization, as in
\citealt{dempster1977maximum}, to assign labels. This yields slightly
better quality in our hand-inspection than labels by majority vote,
presumably because it factors individual worker reliability into the
decision making. In the end, we obtained 4,992 valid labeled
instances: 65.2\% labeled as \emph{condescending} (henceforth,
\emph{positive}), and 34.8\% as \emph{non-condescending} (henceforth,
\emph{negative}).\footnote{There was just one case where \emph{cannot
    decide} was the chosen label; it was in Spanish, so we excluded it
  and added a language classification step to our preprocessing
  pipeline.}

To fully balance the dataset, we pulled out one random month's data
for each year in 2011 to 2017. We extracted instances using the same
methods as described above, but we filtered out \rcomment{}/\rreply{} pairs in
which a condescension-related word appeared. Our final dataset thus
consists of annotated positive and negative instances, with
supplemental randomly-sampled negative instances. For our experiments,
we partitioned the data into 80\% train, 10\% development, and 10\% test
splits. In addition, to simulate real-world situations, we built a
dataset with a 1:20 ratio of positive to negative instances.\footnote{To the best of our knowledge, there is no prior work on what percentage of conversations on Reddit (or, more broadly, in daily conversations) are condescending. Thus, we chose the ratio based on informal observations on Reddit. } The basic statistics of the dataset are shown in Table \ref{tab:corpus-stats-after-augmenting}.

\begin{table}[tp]
  \centering
  \resizebox{0.9\linewidth}{!}{
  \begin{tabular}[t]{r r r}
    \toprule
    & Positive & Negative\\
    \midrule
    Balanced  (1:1) & 3,255 & 3,255\\
    Imbalanced (1:20) & 3,255 & 65,100\\
    \bottomrule
  \end{tabular}
  }
  \caption{Basic statistics of our dataset.
  }
  \label{tab:corpus-stats-after-augmenting}
\end{table}

\section{Experiments}
We now establish some baselines for the \dataset\ Corpus and begin to
test the hypothesis that contextual representations are valuable for
this task. To do this, we use the BERT model of
\citet{devlin2019bert}, which uses a Transformer-based encoder
architecture \citep{Vaswani-etal:2017} to learn word representations
by training against a masked language modeling task and a
next-sentence prediction task. Our models are initialized with the
pretrained representations released by the BERT team and a fully connected layer on the top (Figure 4 in \citealt{devlin2019bert}), which is then
fine-tuned to our dataset \citep{peters2019tune}.  We
explore both BERT Base (\bertb{}) and BERT Large (\bertl{}),\footnote{The whole-word masking model was used as it performs better than the original one in multiple benchmarks.} to
determine whether the added expense of using \bertl{} is justified.
\Sref{appendix-model} provides details on our process of
hyperparameter tuning and optimization.

\begin{table}[tp]
  \centering
  \resizebox{1\linewidth}{!}{
    \begin{tabular}{r @{\hspace*{1.5mm}}c@{\hspace*{1.5mm}} c @{\hspace*{1.5mm}}c@{\hspace*{1.5mm}} l cc}
      \toprule
      Input 1 && Input 2 & & Model &  Imb.~F1  & Bal.~F1\\
      \midrule
      \rquoted{}  & $\wedge$ & \rcontext{} & $\wedge$ & $\bertl{}$ & \textbf{0.684} & \textbf{0.654} \\
      \rquoted{} & $\wedge$ & \rcontext{} & $\wedge$ & $\bertb{}$ & 0.657 & 0.596\\
       \rquoted{}  &  & & $\wedge$ & $\bertl{}$ & 0.650 & 0.640 \\
        &  & \rcontext{}  & $\wedge$ & $\bertl{}$ & 0.611 & 0.513 \\

              &&&& \textit{random} & 0.371 & 0.500 \\
              &&&& \textit{majority} & 0.488 & 0.333 \\
      \bottomrule
    \end{tabular}
  }
  \caption{Performance (macro-F1) for predicting condescension on
    balanced and imbalanced versions of \dataset.
    Model selection was done according to the procedure
    described in \Sref{appendix-model}.}
  \label{tab:perf}
\end{table}

\subsection{Predicting Condescension}

Table~\ref{tab:perf} summarizes the results of our core experiments.
Input~1 and Input~2 describe the basis for the feature
representations. Thus, for example, $\rquoted{} \wedge \rcontext{}$ is
a model that uses both the quoted span and the preceding linguistic
context.  We report two testing scenarios: \emph{Balanced} and
\emph{Imbalanced}, in which there are 20 negative examples for each
positive example.

The results clearly support our hypothesis that context matters; using
the \rquoted{} part and \rcontext{} together give us 3--4\% boost in
macro-F1 using the same model architecture. In addition, we see that
increasing the capacity of the model also helps, though more modestly. It's noteworthy that the performance of using the \rquoted{} part is
better than that of using \rcontext{} alone, though the \rquoted{} part is
roughly three times shorter. Thus, there is a strong signal in the
\rquoted{} part -- the replier chose this span for a reason -- but the
context contains a signal as well.

\begin{table}[tp]
  \centering
  \resizebox{1\linewidth}{!}{%
  \begin{tabular}{ r cc c cc}
    \toprule
    & \multicolumn{2}{c}{\rquoted{} + \rcontext{}} &&  \multicolumn{2}{c}{\rquoted{}}\\
    Ratio & ~~Imb.~F1 & Bal.~F1 && Imb.~F1  & Bal.~F1\\
    \midrule
    1:1   & ~~0.542 & \textbf{0.708} && 0.554 & \textbf{0.690} \\
    1:20  & ~~0.670 & 0.574 && 0.620 & 0.518\\
    2:20  & ~~0.682 & 0.619 && 0.640 & 0.554\\
    3:20  & ~~\textbf{0.684} & 0.654 && 0.646 & 0.585\\
    4:20  & ~~0.678 & 0.632 && \textbf{0.650} & 0.640\\
    5:20  & ~~0.668 & 0.626 && 0.645 & 0.582 \\
    10:20 & ~~0.672 & 0.656 && 0.641 & 0.640 \\
    15:20 & ~~0.665 & 0.641 && 0.641 & 0.593 \\
    20:20 & ~~0.674 & 0.621 && 0.645 & 0.597 \\
    \hline
  \end{tabular}}
\caption{The impact of different train-set positive:negative ratios.
  All the models are \bertl{}. The first row is based on the balanced dataset, and the rest are based on the imbalanced dataset with different oversampling ratios.
  Model selection again used the
  procedure in \Sref{appendix-model}.}
\label{tab:perf-ratio}
\vspace{-1.5mm}
\end{table}

\begin{figure*}[tp]
  \centering
  \includegraphics[width=1\linewidth]{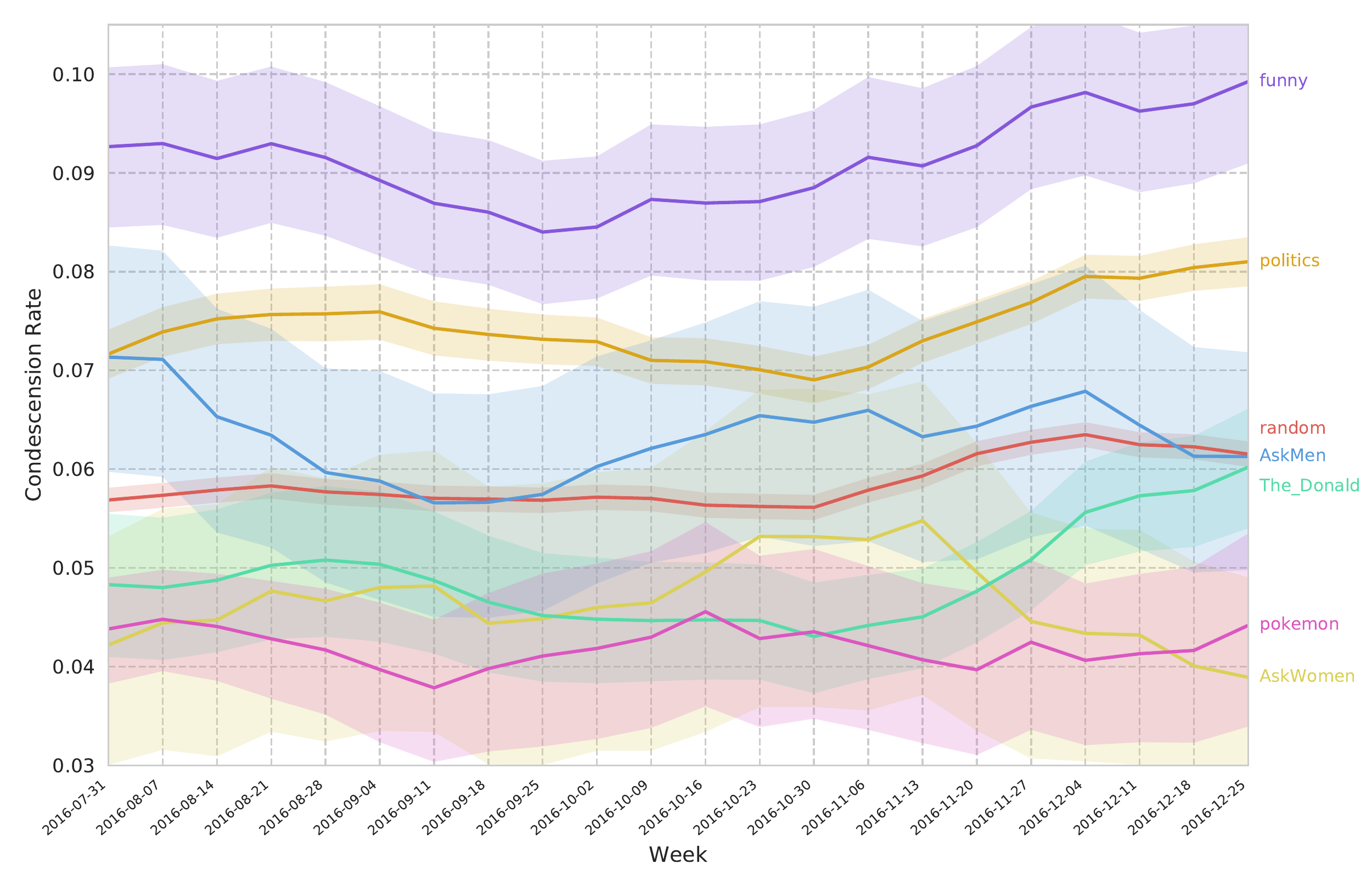}
  \caption{Condescension rates over time in a selection of subreddits,
    as predicted by our best model. The time window is centered around
    the 2016 U.S.\ Presidential Elections. We calculate a rolling mean with
    a window size of 5 and use July 2016 only for smoothing. We obtain 95\% confidence intervals via bootstrapping.}
  \label{fig:experiment}
\end{figure*}

\subsection{Imbalanced Testing Scenarios}

Imbalanced testing scenarios are more challenging, but they also
better reflect usage rates of condescending language in public forums
like Reddit. To further understand how best to get traction on this
problem, we explored a range of different methods for creating
training data. Our results are summarized in
Table~\ref{tab:perf-ratio}. As expected, the balanced problem is best
addressed with a balanced dataset. For the imbalanced problem, we
found that an oversampling ratio of 2 to 4 yielded the best
performance.  Our full \textsc{Quoted} $\wedge$ \textsc{Context} model is
again clearly superior in these scenarios.

\subsection{Condescension Rates Across Subreddits}

Our hope for \dataset\ is that it will play a role in developing
systems that can help identify condescending acts on social media.
This will depend on models trained on \dataset\ being able to
get an accurate read on condescension at scale. As a first step
towards assessing this capability, we ran our models on 14 subreddits, over the time period of July 2016 to December 2016, which covers the 2016 U.S. Presidential Election, an event that we expect to
influence condescension rates in various ways across
Reddit. \Sref{appendix-subreddit} lists these subreddits along with their post
counts and estimated average rates of condescension. Figure~\ref{fig:experiment}
highlights a selection of them. %

As a baseline, we include a 10\% random sample from the top 100 most active subreddits.\footnote{This is derived from the `subscribers' section in {\scriptsize \url{http://redditlist.com/all}}, excluding `announcements'.} Consistently above this baseline are
`politics' and `funny'. It makes sense
that an overtly political subreddit would show a high rate of
condescension (as do `news' and `worldnews'; \Sref{appendix-subreddit}): it's a contentious topic in a contentious time period; see also the rising rate for `The\_Donald' in the post-election period.
It is more surprising that `funny' shows the
highest rates. We do not have a deep understanding of
 why this is, but
it could trace to our model
confusing irony and sarcasm with condescension.

Below the baseline are `AskWomen' and `pokemon'.  We expect `pokemon' to
have low rates of condescension, as it strikes us as a supportive
community. However, one might be surprised to see `AskWomen' so low, especially as compared with `AskMen', which has high rates in general. There is wide support for the idea that women experience
more condescension than men do
\citep{Hall:Braunwald:1981,harris1993provoking,mckechnie1998nature,Cortina-etal:2002,trix2003exploring},
as reflected in the recent lexical innovation \emph{mansplaining},
which can be roughly paraphrased as `a man condescending to a
woman'.\footnote{\scriptsize\url{https://en.wikipedia.org/wiki/Mansplaining}}
However, community norms on `AskWomen' and `AskMen' are likely shaping
these outcomes. Whereas the description for `AskWomen' says it is
``curated to promote respectful and on-topic discussions, and not
serve as a debate subreddit'', the description for `AskMen' ends with ``And
don't be an asshole. Also, go away.''

\section{Conclusion}
We introduced \dataset, a new annotated Reddit corpus of condescending
linguistic acts in context.
Using BERT, we established baseline
models that suggest this is a challenging task, and one that benefits
from rich contextual representations. Finally, in qualitative analyses
on diverse subreddits, we offered initial evidence that models trained
on \dataset\ generalize to new data, a prerequisite for using them to
help improve online communities via condescension detection. The full dataset with the pretrained BERT model is available at {\url{http://github.com/zijwang/talkdown}}.

\section*{Acknowledgements}
We would like to thank David Jurgens for his help with the Reddit dataset, Yulia Tsvetkov and Dan Jurafsky for their insightful early-stage suggestions, Lucy Li for detailed comments on an early draft of the paper, and our
anonymous EMNLP--IJCNLP reviewers for their helpful feedback. We would also like to thank our Mechanical Turk contributors for their help in annotating the \dataset{} Corpus. This research is based in part upon work supported by a Google Faculty Research Award and by the NSF under Grant No. BCS-1456077.

\newpage
\bibliography{main}
\bibliographystyle{acl_natbib}

\newpage

\appendix

\section{Data} \label{appendix-data}
\subsection{In-house Annotation Analysis} \label{appendix-data-inhouse}
\begin{figure}[H]
  \centering
  \includegraphics[width=0.75\linewidth]{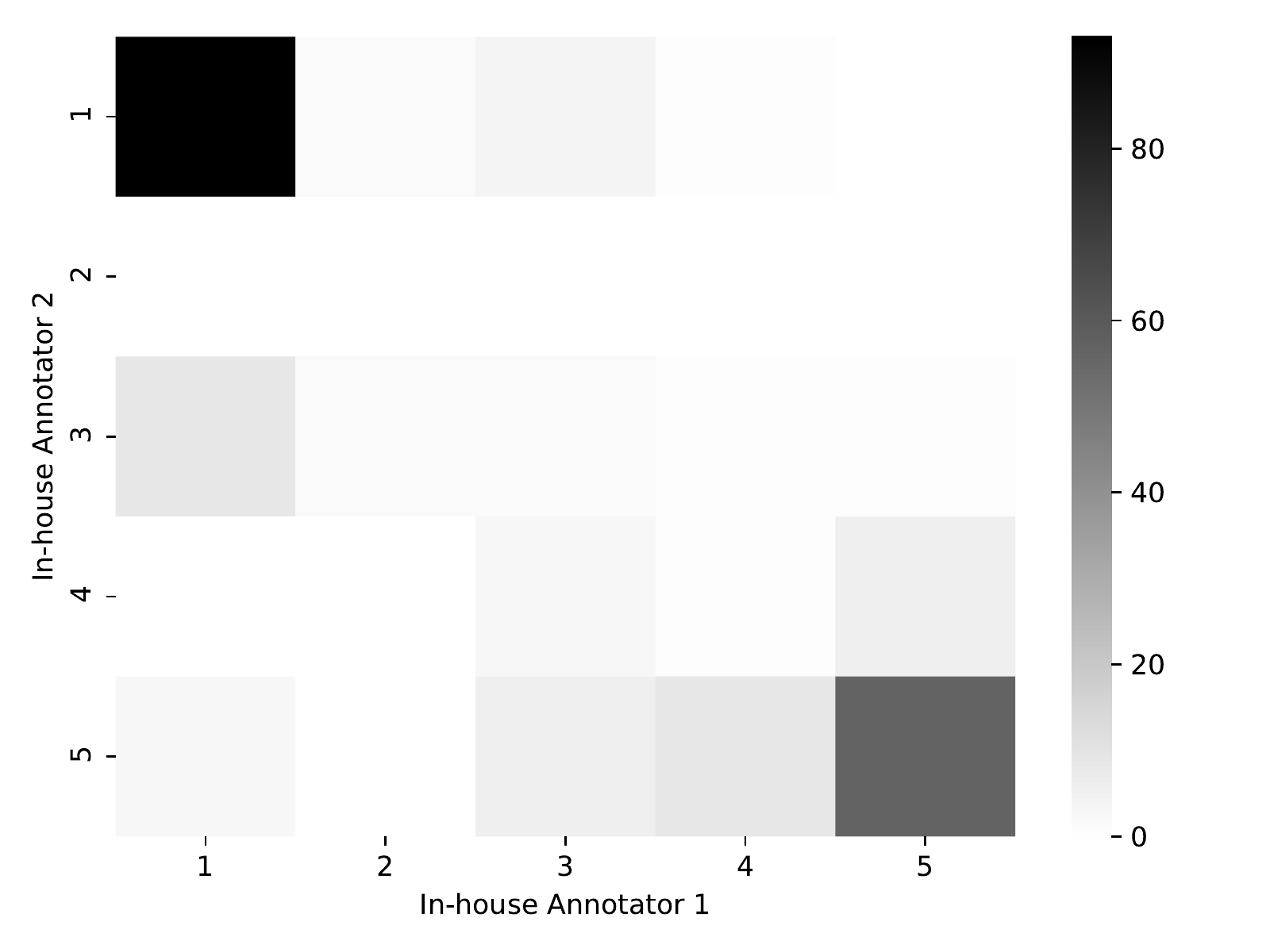}
  \caption{Heatmap for in-house initial assessment with a five-point Likert scale.}
  \label{fig:annotation-inhouse}
\end{figure}

Figure~\ref{fig:annotation-inhouse} shows the five-point Likert scale annotations between two in-house annotators. It can be seen that the signal of condescending or not is clear, and the agreement level between the two annotations is substantial: the Fleiss' $\kappa$ is $0.613$ for the five-point scale and $0.732$ when normalized to three-point scale used in the paper \citep{fleiss1971measuring, landis1977measurement}.

\subsection{Annotation Interface} \label{appendix-data-annotation}
\begin{figure*}[tp]
  \centering
  \frame{\includegraphics[width=1\linewidth]{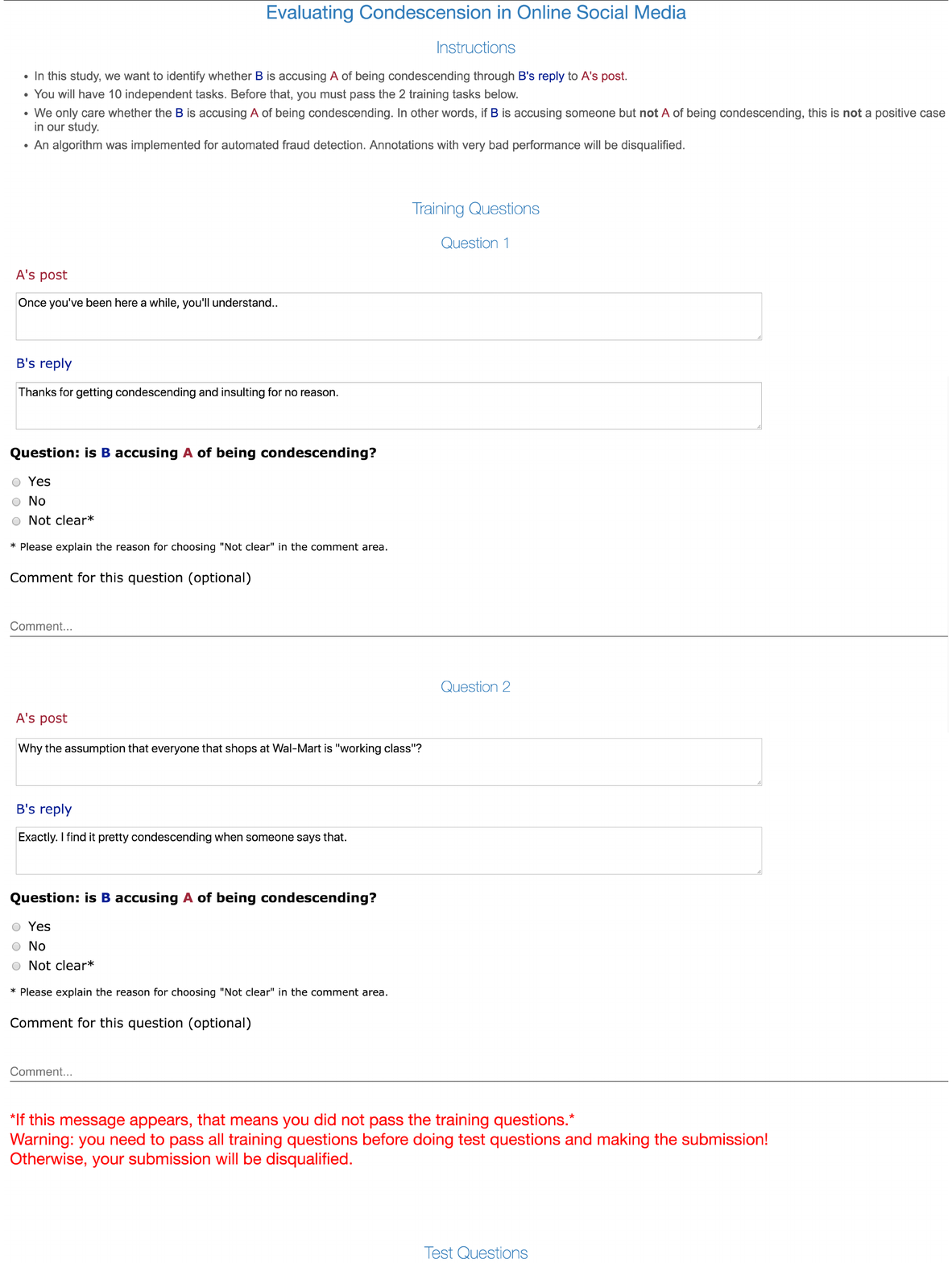}}
  \caption{The initial view of instructions and training questions in the annotation interface.}
  \label{fig:annotation-train}
\end{figure*}

\begin{figure*}[tp]
  \centering
  \frame{\includegraphics[width=1\linewidth]{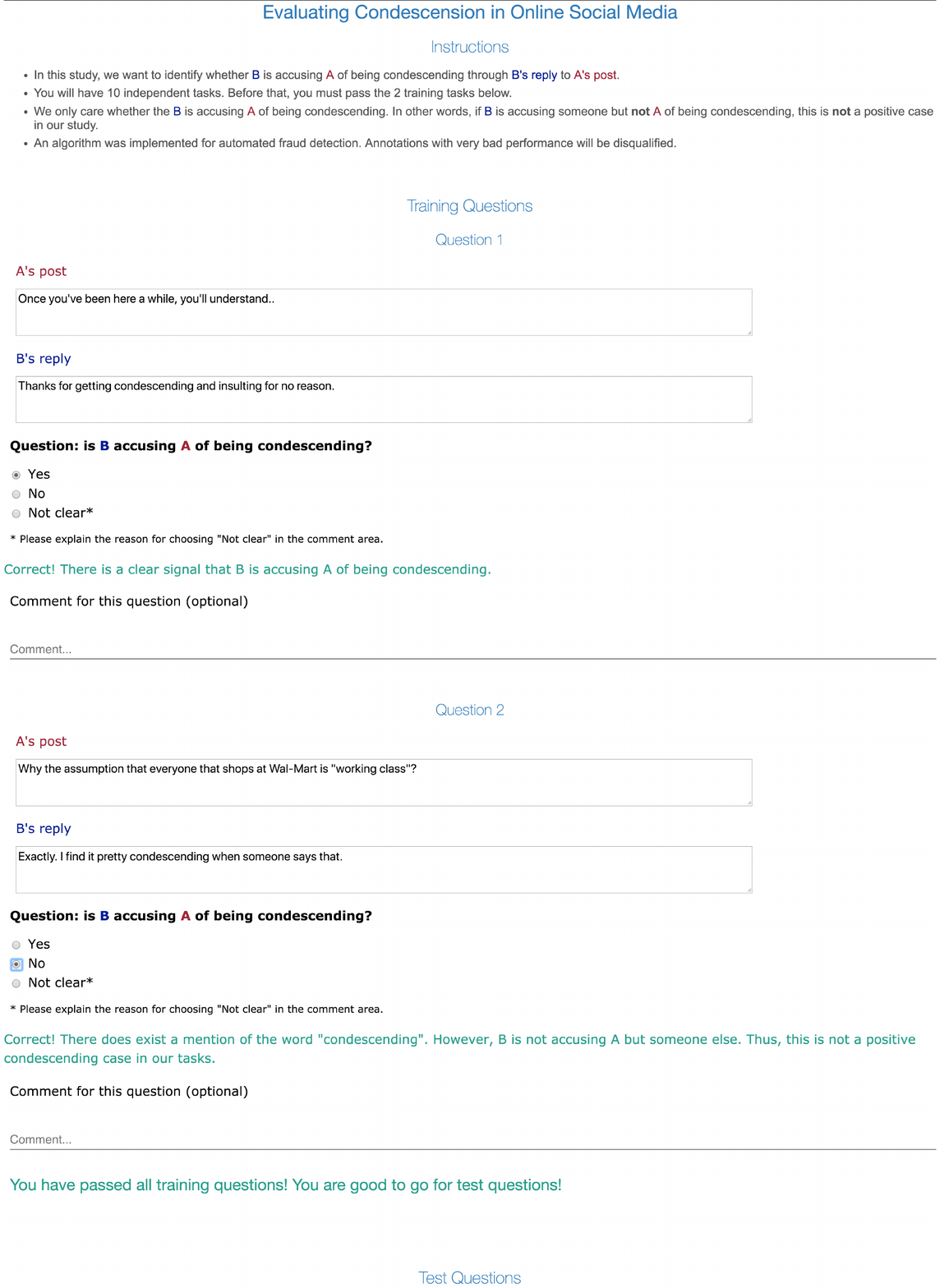}}
  \caption{The view after the annotator passed the training questions.}
  \label{fig:annotation-train-pass}
\end{figure*}

In this section, we show examples of the annotation interface we used on Amazon Mechanical Turk: Figure~\ref{fig:annotation-train} and Figure~\ref{fig:annotation-train-pass}.

Annotators were presented with the task name, the instructions, and two simple training questions, followed by a warning in red saying they needed to pass the training questions to proceed (Figure~\ref{fig:annotation-train}). They had unlimited trials for the training questions, and explanations (for both correct and incorrect answers) were presented directly after each trial. This helped the annotators learn how to approach the task.

After they passed the training questions, they were prompted that they could start to do the test questions (Figure~\ref{fig:annotation-train-pass}). The interface of the test questions %
was similar to that of the training questions, but without explanations after selections. We explicitly checked that the annotators had made selections on each test question before submission, while this was not forced for training questions. This was to filter out possibly low-quality annotations, where the annotators did not pay attention to the instructions.

\section{Model Hyperparameters} \label{appendix-model}
Our BERT models were trained using a set of hyperparameters based on the recommendations in \citealt{devlin2019bert}. Specifically, we set:

\begin{itemize}
\setlength\itemsep{0em}
\item Model Architecture:
\begin{itemize}
\setlength\itemsep{0em}
    \item \bertb{}: Bert Base, Cased
    \item \bertl{}: Bert Large, Cased, with whole-word masking
\end{itemize}
    \item Learning rate: $\{0.5, 0.8, 1, 2, 3, 5\} \cdot 10^{-5}$
    \item Epoch: $2$, $3$
    \item Batch size: $32$
    \item Max sequence length: $512$
\end{itemize}

When optimizing these models, we set the batch size to 32 in order to
ensure there was at least one positive instance per mini-batch. Grid
search was performed with different learning rates and oversampling ratios, and best models were selected based on the best performance on the development set under the imbalanced setting. We found that oversampling 2 to 4 times the positive class
(i.e., 10\%--20\% of the number of instances in the negative class)
generally yielded good performance in all the experiments we ran. For all experiments, we used the HuggingFace PyTorch implementation of BERT.\footnote{\scriptsize\url{https://github.com/huggingface/pytorch-transformers/}}

\newpage
\section{Subreddit Condescension Rates}
\label{appendix-subreddit}

Table~\ref{tab:subr-stats} shows basic statistics for all the subreddits we analyzed.
\begin{table}[H]
\centering
\resizebox{1\linewidth}{!}{%
\begin{tabular}{r r r r}
  \toprule
  Subreddit &   \#Pairs &  Mean rate & Std.~err\\
  \midrule
  \href{https://www.reddit.com/r/4chan/}{4chan} & 6,099 &      0.072 &     0.189 \\
  \href{https://www.reddit.com/r/AskMen/}{AskMen} &19,406 &      0.064 &     0.164 \\
  \href{https://www.reddit.com/r/AskReddit/}{AskReddit} & 256,181 &      0.049 &     0.147 \\
  \href{https://www.reddit.com/r/AskWomen/}{AskWomen} & 9,021 &      0.046 &     0.129 \\
  \href{https://www.reddit.com/r/The_Donald/}{The\_Donald} &57,429 &      0.051 &     0.149 \\
  \href{https://www.reddit.com/r/aww/}{aww} &  8,727 &      0.059 &     0.168 \\
  \href{https://www.reddit.com/r/funny/}{funny} &71,875 &      0.092 &     0.214 \\
  \href{https://www.reddit.com/r/gaming/}{gaming} &64,881 &      0.064 &     0.175 \\
  \href{https://www.reddit.com/r/news/}{news} & 396,710 &      0.066 &     0.179 \\
  \href{https://www.reddit.com/r/pokemon/}{pokemon} & 38,353 &      0.043 &     0.142 \\
  \href{https://www.reddit.com/r/politics/}{politics} &583,033 &      0.075 &     0.186 \\
  \href{https://www.reddit.com/r/stopdrinking/}{stopdrinking} &    910 &      0.038 &     0.112 \\
  \href{https://www.reddit.com/r/tifu/}{tifu} &    35,443 &      0.068 &     0.177  \\
  \href{https://www.reddit.com/r/worldnews/}{worldnews}  &   164,302 &      0.064 &     0.176\\
  \textit{random} &  1,700,192 &      0.059 &     0.166 \\
  
\bottomrule
\end{tabular}}
\caption{Subreddit experiment statistics.
  The raw data are from the Reddit dump from July 2016 to December 2016.
  `Pairs' are \rcomment{}/\rreply{} pairs as defined in the paper.
  `Mean rate' is the mean rate of condescension as estimated by our
  best model, and `Std.~err' gives the associated standard error. `\textit{random}' is a 10\% random sample from the top 100 active subreddits over the same time period.}

  \label{tab:subr-stats}
  \vspace{-1.5mm}
\end{table}

\end{document}